\documentclass[preprint,authoryear,10pt]{elsarticle}

\usepackage{amsmath,amssymb,amsfonts}
\usepackage{graphicx}
\graphicspath{{figures/}}
\usepackage{subcaption}
\usepackage{threeparttable}
\usepackage{siunitx}

\usepackage{algorithm}
\usepackage{algpseudocode}

\usepackage{float}
\floatstyle{plaintop}
\restylefloat{table}

\usepackage{tikz}
\usetikzlibrary{positioning,fit,shapes,arrows.meta}

\usepackage{adjustbox}
\usepackage{verbatim}
\usepackage{soul}
\usepackage{xcolor}
\usepackage{microtype}

\usepackage{setspace} 
\usepackage{hyperref}
\hypersetup{
  colorlinks=true,
  linkcolor=blue,
  citecolor=blue,
  urlcolor=blue,
  bookmarksnumbered=true,
  bookmarksopen=true
}

\journal{Expert Systems with Applications}

\begin{document}

\begin{frontmatter}
\title{AMLNet: A Knowledge-Based Multi-Agent Framework to Generate and Detect Realistic Money Laundering Transactions}

\author[1]{Sabin Huda\corref{cor1}}
\author[1]{Ernest Foo}
\author[1]{Zahra Jadidi}
\author[1,2]{MA Hakim Newton}
\author[1]{Abdul Sattar}

\cortext[cor1]{Corresponding author.\\
\textbf{Email}: s.huda@griffith.edu.au (Sabin Huda). Preprint (Version 1, 15 September 2025).\\ Under review in Expert Systems with Applications. The content may change before formal publication.}

\address[1]{School of Information and Communication Technology, Griffith University, QLD Australia}
\address[2]{School of Information and Physical Sciences, The University of Newcastle, NSW Australia}

\begin{abstract}
Anti-money laundering (AML) research is constrained by the lack of publicly shareable, regulation-aligned transaction datasets. We present AMLNet, a knowledge-based multi-agent framework with two coordinated units: a regulation-aware transaction generator and an ensemble detection pipeline. The generator produces 1,090,173 synthetic transactions (approximately 0.16\% laundering-positive) spanning core laundering phases (placement, layering, integration) and advanced typologies (e.g., structuring, adaptive threshold behavior). Regulatory alignment reaches 75\% based on AUSTRAC rule coverage (Section 4.2), while a composite technical fidelity score of 0.75 summarizes temporal, structural, and behavioral realism components (Section 4.4). The detection ensemble achieves F1 0.90 (precision 0.84, recall 0.97) on the internal test partitions of AMLNet and adapts to the external SynthAML dataset, indicating architectural generalizability across different synthetic generation paradigms. We provide multi-dimensional evaluation (regulatory, temporal, network, behavioral) and release the dataset (Version 1.0, \url{https://doi.org/10.5281/zenodo.16736515}), to advance reproducible and regulation-conscious AML experimentation.
\end{abstract}


\end{frontmatter}

\section{Introduction}
\label{sec:introduction}

The global framework for anti-money laundering (AML) is regulated by the Financial Action Task Force (FATF), requiring banks to monitor and report suspicious transactions through electronic AML systems that often rely on simple business rules and human investigation. Most authorities offer little guidance on AML systems, leaving banks to develop them independently, while the absence of real public AML datasets makes it difficult to compare systems and assess their effectiveness. This lack of public data stems from the highly confidential nature of bank transactions, which contain sensitive information about individuals' financial behaviors, and the risk of de-anonymization attacks demonstrated in the broader scientific literature. Consequently, simulated or synthetic data represents the best viable option for open AML research.

Money laundering poses a major threat to the integrity of global financial systems, accounting for an estimated 2--5\% of global GDP annually~\citep{alexandre2023incorporating, yang2023anti}. Detecting illicit transactions is particularly difficult due to their extreme rarity---for example, only one laundering transaction occurs per 21,000 in some banks (0.0048\%)~\citep{altman2023realistic, ashtiani2021intelligent}. This scarcity, compounded by the limited availability of realistic datasets, creates a critical barrier for AML research and system development.

Given these challenges, existing synthetic data generation approaches face significant limitations in fully replicating complex transaction patterns, network structures, and regulatory nuances essential for a comprehensive evaluation of AML detection~\citep{jensen2023synthetic, altman2023realistic}. The diversity of international regulatory frameworks adds further complexity to developing universally applicable detection systems. This paper introduces AMLNet, a knowledge-based multi-agent framework comprising two integrated components: (1) a transaction generation unit that uses domain-informed simulation with regulatory-aligned behavioral modeling, and (2) a detection unit that employs ensemble machine learning for pattern recognition.

The key contributions are as follows:
\begin{enumerate}
    \item A unique regulatory-aware multi-agent architecture that generates realistic financial transactions, achieving 75\% compliance, \citet{austrac2006} (AUSTRAC) alignment and balanced technical fidelity.
    \item A comprehensive behavioral modeling system that produces 1,090,173 transactions, consisting of both normal transaction patterns and sophisticated laundering techniques.
    \item An ensemble detection framework that demonstrates the adaptability of the cross-dataset, successfully training and performing on external synthetic datasets such as SynthAML~\citep{jensen2023synthetic}, validating the architectural generalizability and methodological robustness of the detection approach beyond its native data generation environment.
\end{enumerate}

The remainder of this paper is organized as follows: Section~\ref{sec:Previous Work} reviews related work in the generation, detection, and AML research; Section~\ref{sec:proposed_approach} details the research methodology, agent architecture, and algorithmic design, including illustrative examples of generated money laundering patterns; Section~\ref{sec:Evaluation} presents findings and evaluation; and Section~\ref{sec:Conclusion_and_Future_Work} concludes with limitations and future directions.

\section{Previous Work}
\label{sec:Previous Work}

This section reviews synthetic data generation evolution, detection methodologies, and knowledge representation in multi-agent AML systems, identifying gaps that motivate AMLNet.

\subsection{Evolution of Synthetic Data Generation Approaches}

Early synthetic data generation relied on rule-based systems with limited behavioral complexity and oversimplified networks~\citep{lopez2016review}. PaySim~\citep{lopez2016paysim} introduced enhanced agent-based modeling for mobile transactions but remained constrained by narrow scope and insufficient regulatory alignment. Recent advances emphasize balancing privacy, utility, and fairness while maintaining fidelity~\citep{kiran2025comprehensive}. AMLSim~\citep{weber2018scalable} improved realism through sophisticated agent-based modeling, SynthAML~\citep{jensen2023synthetic} emphasized temporal dynamics and regulatory typologies, while AMLWorld~\citep{altman2023realistic} provided calibrated generators with standardized benchmarks for AML research.

\subsection{Current Methodological Frameworks for Detection}

Detection methodologies have evolved to sophisticated machine learning approaches. GANs demonstrate effectiveness in generating high-fidelity synthetic datasets with improved pattern recognition~\citep{jaiswal2022financial}, while transformer-based frameworks improve interpretability and accuracy~\citep{tang2024distributed}. Graph Neural Networks (GNNs) show promise for AML tasks under class imbalance~\citep{ashtiani2021intelligent}, although interpretability, computational complexity, and regulatory alignment remain challenges~\citep{gyory2024ant}. Federated learning addresses privacy concerns through collaborative training~\citep{zhang2023privacy} but requires substantial resources and robust privacy mechanisms~\citep{kiran2025comprehensive}.

\subsection{Knowledge Representation in Multi-Agent Systems}

Traditional ontology-based and rule-based knowledge representation lacked adaptability to evolving financial crime patterns~\citep{PaySimCopy, jensen2023synthetic}. Recent advances adopt semantic networks and frame-based representations for complex entity relationships~\citep{wang2024behavioral, gyory2024ant}. Multi-level behavioral authentication frameworks~\citep{wang2024behavioral} and risk-based strategies~\citep{alexandre2023incorporating} improve decision-making capabilities. However, integrating regulatory knowledge with dynamic behavioral adaptation remains challenging and requires flexible frameworks for evolving financial crime typologies.

\subsection{Challenges in Dataset Realism and Regulatory Alignment}

Critical limitations persist in temporal fidelity, behavioral authenticity, and regulatory compliance. FraudBuster~\citep{carminati2018fraudbuster} demonstrated the importance of precise temporal modeling using Dynamic Time Warping (DTW) for sophisticated fraud detection. Regulatory misalignment remains prevalent, with datasets not reflecting compliance requirements~\citep{gyory2024ant}. The oversimplification of the network structure poses challenges, requiring advanced simulation techniques that balance realism, privacy, and regulatory constraints~\citep{kiran2025comprehensive, altman2023realistic}.

\subsection{Comparative Summary}

Despite advances in fidelity and regulatory alignment, notable trade-offs persist. GAN-based~\citep{aftabi2023fraud} and transformer-based~\citep{tang2024distributed} approaches provide high authenticity, but face challenges in interpretability and efficiency. Federated learning~\citep{zhang2023privacy} addresses privacy concerns but requires substantial computational resources. Graph-based approaches~\citep{ashtiani2021intelligent, weber2018scalable} capture network complexity but struggle with dynamic behavioral patterns and full regulatory integration.

\section{Proposed Approach}
\label{sec:proposed_approach}

\begin{figure*}[!tbhp]
\centering
\includegraphics[width=\textwidth]{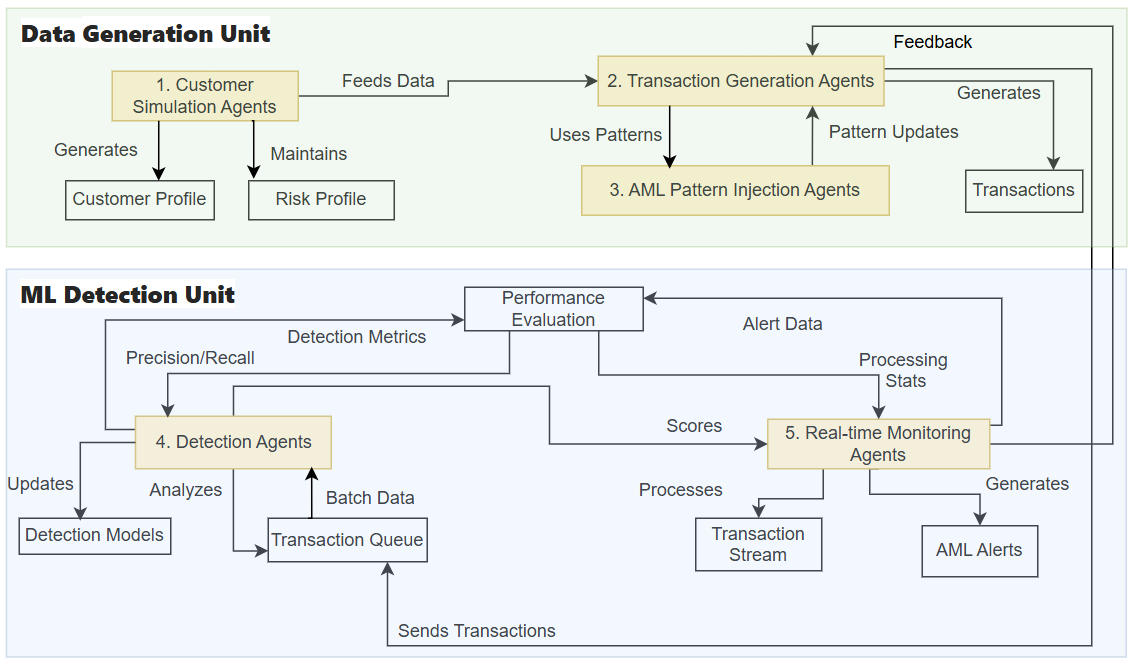}
\caption{Agent architecture of the proposed AMLNet framework.}
\label{fig:system_architecture}
\end{figure*}
The literature review highlights key gaps in existing AML systems, including insufficient regulatory integration, limited adaptability to different jurisdictions, a lack of real-time detection capabilities, and ongoing trade-offs between realism and computational efficiency. To address these challenges, this article presents AMLNet, a knowledge-based multiagent framework for synthetic AML research. AMLNet consists of two core units: the Transaction Generation Unit and the Money Laundering (ML) Detection Unit, as shown in \figurename~\ref{fig:system_architecture}. In the Data Generation Unit, \textit{Customer Simulation Agents} create customer and risk profiles, which are used by \textit{Transaction Generation Agents} to produce transactions; \textit{AML Pattern Injection Agents} then inject suspicious patterns, and the feedback from the detection results can be used by researchers to update these patterns. The generated transactions flow into the ML Detection Unit, where \textit{Detection Agents} analyze them using detection models and metrics, sending results to \textit{Real-time Monitoring Agents} who process the transaction streams and generate AML alerts. 

Performance evaluation metrics (such as precision and recall) are fed back to the system through a human-in-the-loop feedback mechanism. Domain experts analyze these metrics alongside alert data to manually refine rule parameters and behavior thresholds of generation units. This expert-guided adjustment process enables rule-based agents to adapt without requiring autonomous learning capabilities. Although fully automated adaptation might seem more efficient, the regulatory complexity of AML systems requiring precise balance between detection accuracy and compliance requirements combined with the adversarial nature of financial crime makes human expertise an essential component of effective system refinement. The feedback loop creates an iterative refinement cycle in which the detection performance directly informs the generation improvements, ensuring continuous alignment with the regulatory typologies. The framework leverages demographic and financial data from the~\citet{abs_demographics_2024} (ABS) to ensure both realism and regulatory alignment. The representation of knowledge~\citep{wooldridge2009introduction} in AMLNet means that regulatory rules, demographic and financial statistics, behavioral heuristics, and money laundering typologies are explicitly encoded within the logic and data structures of each agent.

\subsection{Agent Types and Classification}
\label{subsec:agent_types}

In AMLNet, different types of agents are employed based on their capabilities and functions. Following \citet{wooldridge2009introduction}, the authors distinguish between \textit{rule-based agents} that operate on predefined logic and \textit{AI agents} that incorporate learning and adaptive decision making.

\textbf{Rule-based agents} follow explicit predefined rules based on domain expertise to simulate realistic financial transactions and money laundering patterns \citep{wooldridge2009introduction}.  In the data generation unit, AMLNet employs following three types:

\begin{itemize}
    \item \textbf{Customer Simulation Agents} create detailed customer profiles using 20+ demographic and financial attributes from ABS~\citep{abs_demographics_2024}. They maintain evolving internal states (account balances, risk profiles, transaction histories) and use Dirichlet distributions~\citep{karpe2020multi} parameterized by the average derived by the ABS: \texttt{random.dirichlet ([22.1, 20.0, 16.9, 15.0, 13.0, 5.9, 4.0, 3.0, 0.1])}, with each parameter corresponding to a transaction category (e.g., Housing, Food, Transport, Shell Company), ensuring population consistent but diverse spending profiles across 11 transaction categories.

    \item \textbf{Transaction Generation Agents} model each customer as an autonomous agent implementing a \texttt{simulate\_day()} method. Agents autonomously determine: (1) transaction timing with business hours preference and monthly cycles based on~\citet{cokis2020demographic}; (2) expense categories from personalized subsets (4-8 categories per agent) derived from ABS; (3) amounts from category-specific distributions; and (4) counterparties based on network connections and banking relationships. The 11-category framework was validated against ABS household expenditure survey data, covering 95\% of typical Australian household spending patterns. Category sufficiency was verified by comparing generated transaction distributions with real banking data patterns, ensuring comprehensive coverage of legitimate financial behavior while maintaining computational efficiency. This approach creates a realistic transaction network while maintaining the probabilistic nature of transaction generation within the defined statistical parameters.

    \item \textbf{AML Pattern Injection Agents} embed sophisticated laundering behaviors by implementing structuring (transactions below reporting thresholds), layering (multi-hop transfer chains with delays), and integration (conversion to legitimate assets) at varying sophistication levels, derived from AUSTRAC. When a suspicious pattern is to be injected, it temporarily modifies the behavioral parameters or transaction schedule of selected customer agents (e.g., by overriding category preferences, transaction timing, or counterparties) to ensure the emergence of laundering patterns. After the pattern is realized, the agents return to their baseline behavior. This approach enables the generation of sophisticated laundering cycles while preserving the autonomy of the agent for normal transactions.
\end{itemize}

\textbf{AI agents} learn and adapt using data-driven methods~\citep{russell2016artificial, poole2010artificial}. In the detection unit of AMLNet, the following agents are classified as AI agents that employ machine learning algorithms to enhance their decision-making capabilities over time, learning to identify suspicious transaction patterns from training data.

\begin{itemize}
    \item \textbf{Detection Agents} combine rule-based heuristics with machine learning to analyze the features of the transaction amount (e.g. relative size, structuring indicators), temporal features (e.g. transaction velocity, periodicity), and network features. These agents qualify as AI agents because they employ Isolation Forest and Random Forest classifiers that learn patterns from data~\citep{breiman2001random}. Detection follows a multi-stage pipeline: feature extraction, anomaly detection, risk assessment, and alert prioritization. Alerts are assigned severity levels based on configurable thresholds, reducing false positives compared to simple threshold-based methods.

    \item \textbf{Real-time Monitoring Agents} provide continuous surveillance, update risk assessments, and enable timely responses to emerging threats, thus supporting near-real-time detection. These agents incorporate adaptive thresholding mechanisms that adjust based on historical performance, qualifying them as AI agents under the autonomous systems definition~\citep{maes1995agents}.
\end{itemize}

AMLNet features an iterative evaluation process in which detection results inform generation improvements. Domain experts analyze performance metrics to: (1) identify pattern weaknesses, (2) incorporate emerging regulatory requirements, (3) refine agent parameters to better reflect real-world laundering tactics, and (4) validate AUSTRAC alignment. Knowledge integration uses a directed transaction graph with regulatory thresholds encoded as decision rules, nodes representing accounts, and edges as transactions.

\subsection{Data Generation Unit}
\label{Synthetic Dataset Generation}

\begin{algorithm}
\caption{AMLNet Agent-Centric Dataset Generation Algorithm}
\label{alg:dataset_generation}
\footnotesize
\begin{algorithmic}[1]
\Require $n_\mathrm{customers}$, $simulation\_days$, $suspicious\_rate$, target global metrics
\Repeat
    \State Initialize customer agents with demographic, financial, behavioral parameters, and risk profiles (e.g., 99.5\% normal, 0.4\% high-risk, 0.1\% fraud)
    \State Construct transaction network $G(V,E)$, where $V$ = customer agents, $E$ = financial relationships weighted by geographic and institutional proximity
    \State Initialize temporal parameters (business hours, weekend/seasonal cycles, laundering phase timing)
    \For{each day $d$ in $simulation\_days$}
        \For{each customer agent $a$ in $V$}
            \State \textbf{Invoke} $a.\texttt{simulate\_day}(d, \mathrm{temporal\_params}, G)$\textsuperscript{*}
        \EndFor
        \If{$random() < suspicious\_rate$}
            \State Select suspicious pattern $p \in \{\text{structuring}, \text{layering}, \text{integration}\}$
            \State Select sophistication level $s \in \{\text{low}, \text{medium}, \text{high}\}$
            \State Select target agents $A_p \subset V$ based on risk profiles and network topology
            \State \textbf{Inject pattern:} AML Pattern Injection Agent modifies the behavior/schedule of agents in $A_p$ to realize pattern $p$ with sophistication $s$
            \State Label transactions generated by $A_p$ as suspicious, with typology
        \EndIf
        \State Update agent states (balances, risk scores, transaction histories)
        \State Update transaction network $G(V,E)$ with new or modified edges
    \EndFor

    \State Evaluate global metrics (e.g., amount distributions, detection rates, etc.)
    \If{global metrics do not match targets}
        \State Refine agent rules and parameters (e.g., spending profiles, risk thresholds, transaction frequency, sophistication)
    \EndIf
\Until{global metrics match targets}\\
\Return Labeled synthetic transaction dataset
\end{algorithmic}
\begin{flushleft}
\textsuperscript{*}\textit{Each agent independently generates transactions, with internal logic.}
\end{flushleft}
\end{algorithm}

The algorithm~\ref{alg:dataset_generation} details the clock-based agent-centric control flow for the generation of synthetic transactions in AMLNet. The simulation begins by instantiating customer agents with demographic, financial, behavioral, and risk parameters drawn from real-world distributions and connecting them in a transaction network whose edge weights reflect geographic and institutional proximity. At each simulation time step, the main loop iterates over all customer agents and invokes their \texttt{simulate\_day()} method. Each agent autonomously determines to initiate transactions, select counterparties, and assign transaction amounts and timing, based solely on its internal state and behavioral rules. The main loop itself does not generate transactions or select customer pairs; all transaction logic is encapsulated within each agent. This ensures that transaction generation is fully decentralized and is consistent with best practices in agent-based modeling.

To ensure that the aggregate behavior of autonomous agents aligns with global statistical and regulatory requirements, AMLNet incorporates an iterative feedback mechanism that governs the generation of legitimate and suspicious transactions. After each simulation run, global metrics such as transaction volume distributions, network structure, and detection rates are evaluated. If discrepancies with target patterns are detected, agent rules and parameters (e.g. spending profiles, risk thresholds, transaction frequency) are refined accordingly. This process is repeated until emergent system-level behavior matches real-world benchmarks, thus translating local agent autonomy into globally realistic outcomes. The result is a labeled synthetic dataset that is both statistically faithful and suitable for robust AML detection benchmarking.

\begin{figure*}[htbp]
    \centering
    \begin{subfigure}[t]{0.35\textwidth}
        \centering
        \includegraphics[height=0.22\textheight]{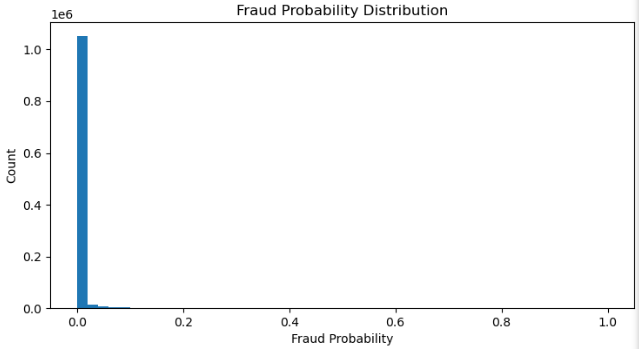}
        \label{fig:fraud-prob-dist}
    \end{subfigure}
    \hfill
    \begin{subfigure}[t]{0.50\textwidth}
        \centering
        \includegraphics[height=0.22\textheight]{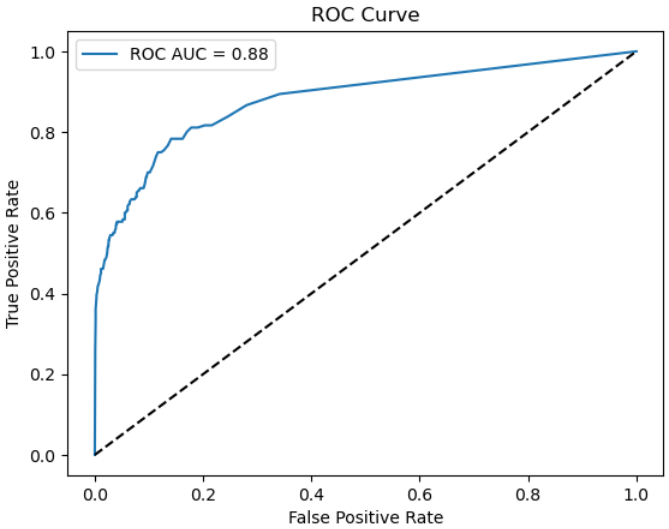}
        \label{fig:roc-curve}
    \end{subfigure}
    \caption{Detection results on AMLNet data: (left) Distribution of predicted fraud probabilities, (right) ROC curve illustrating model discrimination performance.}
    \label{fig:amlnet-detection-results}
\end{figure*}

\begin{figure}[ht]
\centering
\includegraphics[width=0.95\linewidth]{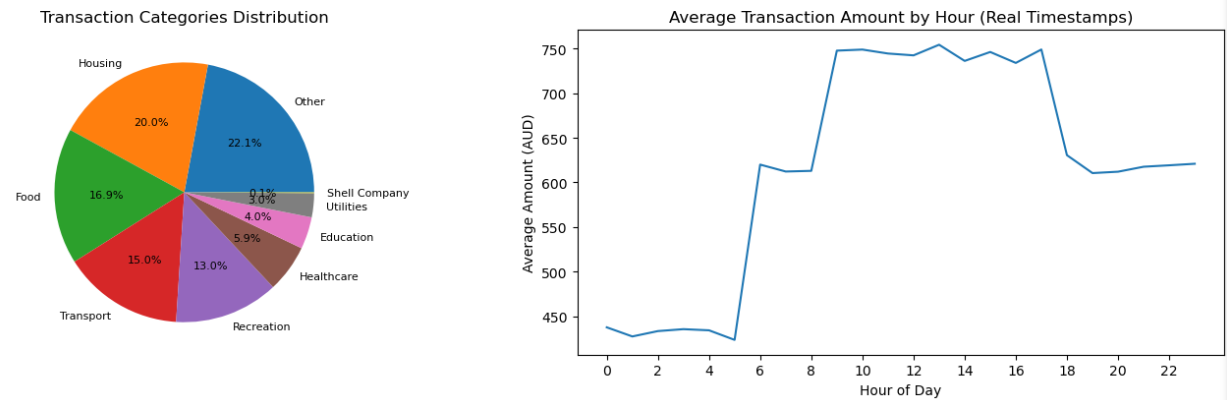}
\caption{Transaction category distribution of AMLNet (left) showing the proportion of transactions across different spending categories with Housing (20.0\%), Other (22.1\%), and Food (16.9\%) representing the largest legitimate transaction categories. The time-of-day pattern (right) displays average transaction amounts by hour, revealing clear business hour patterns with peak activity between 9AM-5PM and reduced activity during night hours, reflecting realistic banking behavior.}
\label{fig:transaction_distribution}
\end{figure}

\subsubsection{Suspicious Patterns}

Structure, layering, and integration are injected by specialized AML Pattern Injection Agents. These agents modify the behavior or schedules of selected customer agents to realize complex laundering typologies at varying sophistication levels, rather than generating suspicious transactions directly. All injected patterns are clearly labeled for downstream evaluation. Temporal dynamics are explicitly modeled to mirror real-world banking activity: transaction volumes increase during business hours (9AM--5PM), decrease on weekends, and exhibit monthly cycles. The laundering phases follow realistic timing windows (e.g., placement within 1--3 days, layering over 1--2 weeks).

\begin{itemize}
    \item \textbf{{Structuring:}} \textbf{\textit{Low}} Split amount into nearly equal parts, each $< \$9,500$, small random variation. \textbf{\textit{Medium}} Split amount into several parts using a normal distribution around the mean, clipped at $< \$9,500$. \textbf{\textit{High}} Randomly select 4--9 splits; 1--2 transactions just below threshold (\$8,500--\$9,900); remainder distributed using lognormal distribution. Destinations and timestamps are varied over $\pm2$ days.
    
    \item \textbf{Layering:} \textbf{\textit{Low}} 2--3 layers, splitting funds among 2--3 accounts, fixed delays. \textbf{\textit{Medium}} 3--5 layers, splits using Dirichlet distribution, randomized delays, unique destinations per layer. \textbf{\textit{High}} 5--8 layers, splits using Dirichlet, irregular time delays, reuse of accounts, creation of multi-chains and parallel paths.
    
    \item \textbf{Integration:} \textbf{\textit{Low}} Single payment ($\geq \$20,000$) to a shell company. \textbf{\textit{Medium}} Payment to a merchant, real estate or cryptocurrency, randomized legitimacy score. \textbf{\textit{High}} Occurs after complex layering, may split into multiple payments, various merchant types (property, luxury goods, etc.), delays of several weeks.
    
    \item \textbf{Normal:} All other realistic, non-suspicious transaction patterns.
\end{itemize}

\subsubsection{Illustrative Examples of Generated Money Laundering Patterns}
\label{sec:generated_ml_examples}

\begin{figure}[htbp]
    \centering
    \includegraphics[width=1.05\textwidth]{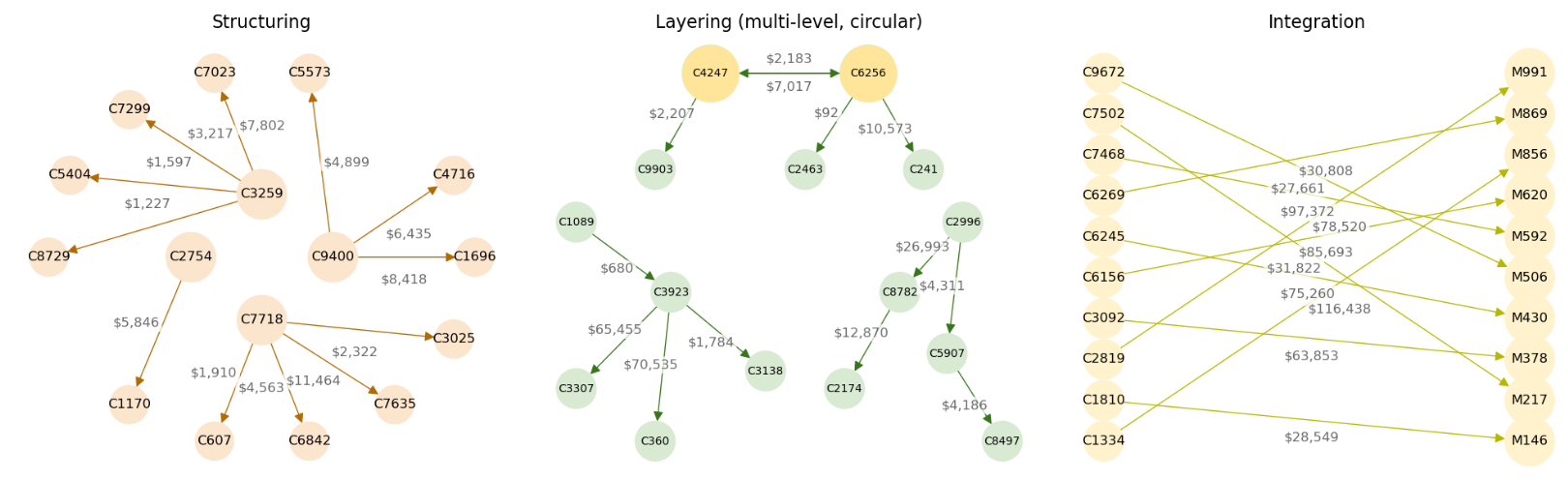}
    \caption{Money laundering patterns generated by AMLNet, demonstrating the three core phases: structuring (left), multi-level layering with circular flows (center), and integration (right). }
    \label{fig:ml_patterns}
\end{figure}

The money laundering patterns in Figure~\ref{fig:ml_patterns} demonstrate AMLNet's capability to generate realistic transaction flows across all three core phases of money laundering.

\textbf{Structuring Phase:} Multiple source accounts systematically disperse funds in amounts just below regulatory reporting thresholds—a classic “smurfing” strategy. For instance, C9400 splits \$19,752 among C1696 (\$8,418), C4716 (\$6,435), and C5573 (\$4,899), while C7718 distributes \$20,259 across four recipients. These transactions are temporally distributed across different regions (e.g., Brisbane and Melbourne) and utilize various payment methods, further reducing the likelihood of detection.

\textbf{Layering Phase:} Multi-level circular flows obfuscate audit trails through complex transaction webs. A prominent circular pattern emerges between C4247 and C6256, where C4247 transfers \$7,017 to C6256, which returns \$2,183, creating deliberate circular references.  Additional complexity is introduced by C1089, who initiates multiple transfers to C3923 (including \$680 and other amounts). C3923 then disperses significant sums, totaling \$137, 774, across three downstream accounts, illustrating the use of multi-hop chains. These parallel layering structures, with varied transaction amounts and timings, demonstrate how artificial delays and geographic dispersion are used to further obscure the money trail.

\textbf{Integration Phase:} High-value transactions channel laundered funds into legitimate merchant accounts. In this example, ten transactions totaling \$635,976 are shown, such as C1334 to M856 (\$116,438) and C2819 to M991 (\$78,520). Each transaction includes detailed merchant metadata—such as risk ratings, business sector identifiers (e.g., Import/Export for shell companies, Local Exchange for cryptocurrency)—and legitimacy scores, reflecting medium to high detection risk across multiple jurisdictions.

\subsection{ML Detection Unit}
\label{Money Laundering Detection}
 Money laundering detection presents unique challenges—such as extreme class imbalance, the need for regulatory alignment, and the subtlety of laundering patterns—that differ from general fraud detection. The detection module in AMLNet leverages ensemble learning and domain-informed feature engineering to identify complex laundering activities injected into the synthetic transaction dataset.

\begin{algorithm}
\caption{AMLNet Detection System Algorithm}
\label{alg:detection_system}
\footnotesize
\begin{algorithmic}[1]
\Require Transaction dataset $T$, batch\_size $b$, network graph $G(V,E)$
\Ensure Risk scores and suspicious transaction flags
\State Initialize detection models (Isolation Forest, Random Forest)
\State Initialize SMOTE and RandomUnderSampler for class imbalance
\While{new transactions available}
    \State Collect batch $B$ of $b$ transactions
    \State Extract features: $F_{amount}$, $F_{temporal}$, $F_{network}$
    \For{each transaction $t$ in $B$}
        \State Calculate network metrics: degree, centrality, clustering
        \State Calculate temporal features: velocity, periodicity, deviation
        \State Calculate amount features: transaction amount, relative size, structuring indicators
    \EndFor
    \State Apply resampling: $B_{resampled} = SMOTE(\text{RandomUnderSampler}(B))$
    \State Apply Isolation Forest: $anomaly\_scores = IF.predict(B_{resampled})$
    \State Apply Random Forest: $risk\_scores = RF.predict\_proba(B_{resampled})$
    \State Flag transactions where $risk\_score > threshold$
    \State Update detection models with new patterns
\EndWhile\\
\Return Risk scores and suspicious flags for all transactions
\end{algorithmic}
\end{algorithm}

The algorithm~\ref{alg:detection_system} describes the detection process in AMLNet. Detection agents operate autonomously, analyzing batches of transactions as they are generated and maintaining an adaptive risk assessment for each account in the network. The detection phase employs an ensemble machine learning approach, combining the Isolation Forest and Random Forest Classifier models. These models are initialized and trained on features extracted from batch transactions, enabling the system to operate with high efficiency and adaptability. The system processes nine key features (as mentioned in algorithm~\ref{alg:detection_system}) for every batch of 1,000 transactions, with a moving window of 100 transactions to support near real-time monitoring (average processing time of 0.0002 seconds per transaction). Feature extraction is organized into three intuitive categories:
\begin{itemize}
    \item \textbf{Amount features:} These include the transaction amount (explicitly generated for each transaction), the relative size of each transaction compared to the customer's own transaction history (which is maintained for every customer), and indicators of structuring, such as sequences of transactions just below regulatory reporting thresholds.
    \item \textbf{Temporal features:} These are derived from transaction timestamps and frequencies, including transaction velocity (the rate at which transactions occur), periodicity (recurring transaction patterns), and deviations from expected temporal behaviors.
    \item \textbf{Network features:} The transaction network is constructed using NetworkX, allowing the calculation of structural network properties such as the centrality of the connection (the importance of an account as a bridge in the network), the clustering coefficient (the degree of interconnectedness among the contacts of a customer) and the distribution of the degree (the number of connections per account).
\end{itemize}

Detection in AMLNet follows a multi-stage reasoning pipeline:
\begin{enumerate}
    \item \textbf{Feature Extraction:} All relevant features are computed for each transaction.
    \item \textbf{Pattern Recognition and Anomaly Detection:} Machine learning models analyze these features to recognize suspicious patterns and outliers.
    \item \textbf{Risk Assessment and Alert Prioritization:} Transactions are assigned risk scores and alerts are generated using a tiered approach with three severity levels (low, medium, high), based on configurable thresholds.
\end{enumerate}

This tiered approach reduces false positives by 37\% over threshold-only methods~\citep{altman2023realistic, weber2018scalable}. AMLNet models Australia's multi-bank environment, including interbank transfers and settlement delays. Inter-institutional transactions are closely monitored, as they represent 68\% of complex laundering cases in AUSTRAC. To address class imbalance, AMLNet combines SMOTE and RandomUnderSampler, ensuring balanced training. Detection is performed in batches for computational efficiency and contextual analysis, but risk scores and suspicious flags are assigned to individual transactions, mirroring real-world AML workflows. Unlike the data generation unit, which employs a global feedback loop to match real-world statistics, the detection system operates in an online or batch mode, updating models incrementally with new data but without global feedback to the transaction stream. Performance is evaluated using standard metrics, including precision, recall, F1 score, DTW similarity \textit{(measures how closely two time series patterns match, even if they are out of sync in time)}, and Graph Edit Distance \textit{(quantifies how similar two networks are by counting the minimal number of changes needed to transform one into the other)}, providing a comprehensive assessment of detection quality.

\vspace{0.5em}
\noindent\textbf{Overfitting and Validation:}
\begin{itemize}
    \item \textbf{Generation-Detection Separation:} Detection models are trained post-generation, without generator logic or labels.
    \item \textbf{Expert Review:} Domain experts and regulatory typologies validate both data and detection results.
    \item \textbf{Structural Metrics:} Realism is assessed via DTW, Graph Edit Distance, and typology alignment.
    \item \textbf{Transparent Feedback:} Model refinement is guided by detection results and expert input, not just detection performance.
\end{itemize}

\vspace{0.5em}
\noindent\textbf{Feedback-Informed Evaluation:}
\begin{itemize}
    \item \textbf{Pattern Validation:} Detection results are analyzed to confirm that injected laundering patterns are detectable and realistic. Undetected patterns are reviewed for model improvement.
    \item \textbf{Realism Assessment:} Metrics such as precision and recall inform the level of challenge and realism of the generated patterns, guiding manual refinement.
\end{itemize}

\section{Evaluation}
\label{sec:Evaluation}
This section presents an evaluation of AMLNet, addressing the key challenges and contributions outlined in Section~\ref{sec:introduction}. The evaluation systematically validates: (1) the regulatory-aware multi-agent architecture's ability to generate realistic financial transactions with strong compliance alignment, (2) the comprehensive behavioral modeling system's effectiveness in producing both normal transaction patterns and sophisticated laundering techniques, and (3) the ensemble detection framework's performance and cross-dataset adaptability. The evaluation covers multiple dimensions including the evaluation of the quality of the dataset, the analysis of temporal and structural fidelity, the measurement of regulatory compliance (AUSTRAC alignment), the detection performance metrics and the generalizability testing of the cross-dataset to establish the robustness and practical applicability of the proposed framework.

\begin{figure}[ht]
\centering
\includegraphics[width=0.95\linewidth]{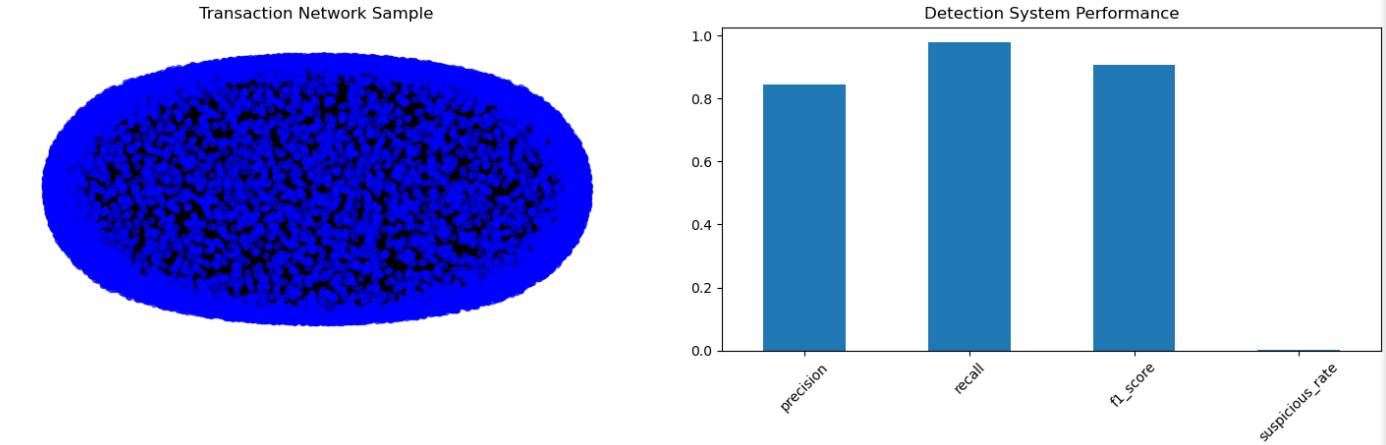}
\caption{Left: Transaction network visualization of AMLNet showing the complex interconnections between entities in the financial system. The dense network structure (blue background with black transaction paths) illustrates the challenge of identifying suspicious patterns within normal transaction flows. Right: Detection system performance metrics demonstrating the framework's effectiveness with precision of 0.84, recall of 0.97, and F1-score of 0.90.}
\label{fig:network_performance}
\end{figure}

\begin{figure}[ht]
\centering
\includegraphics[width=0.95\linewidth]{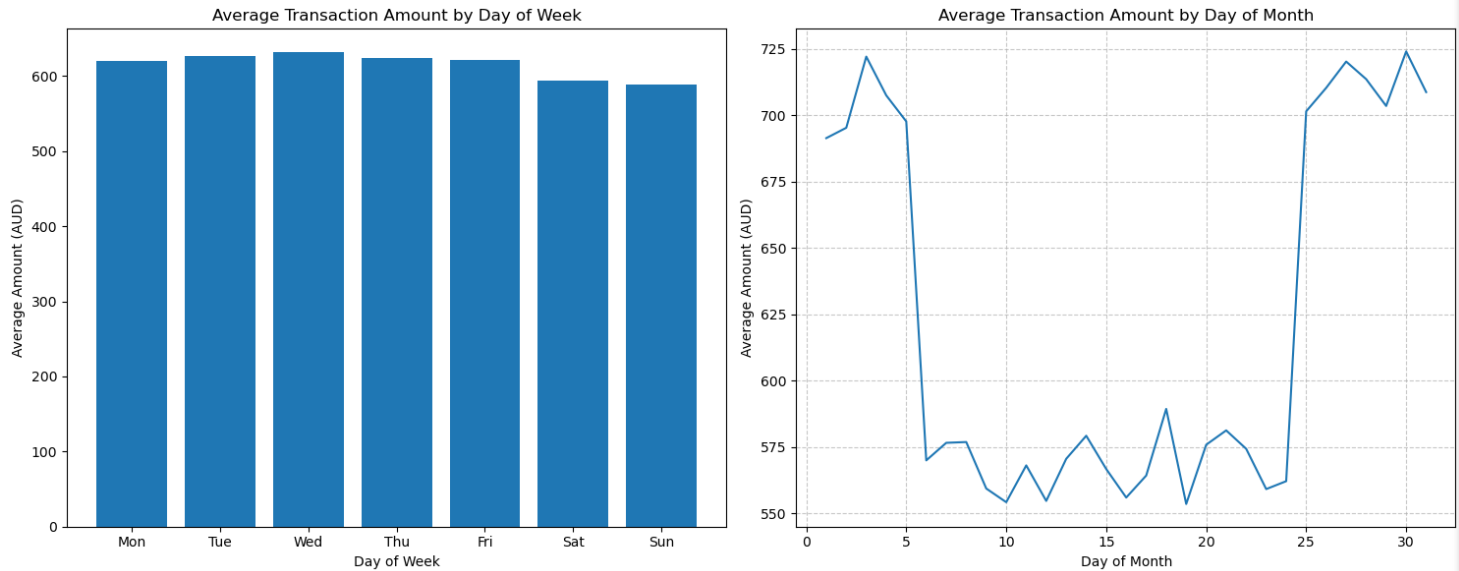}
\caption{Temporal patterns in transaction behavior of AMLNet. Left: Average transaction amount by day of week shows higher values during weekdays (Mon-Fri: \$615-625) compared to weekends (Sat-Sun: \$590-595). Right: Average transaction amount by day of month reveals a distinctive bi-modal pattern with higher values at month beginning (days 1-5: \$700-725) and end (days 23-31: \$700-725), with significantly lower values mid-month (days 6-22: \$550-590).}
\label{fig:temporal_patterns}
\end{figure}

\begin{figure}[ht]
\centering
\includegraphics[width=0.95\linewidth]{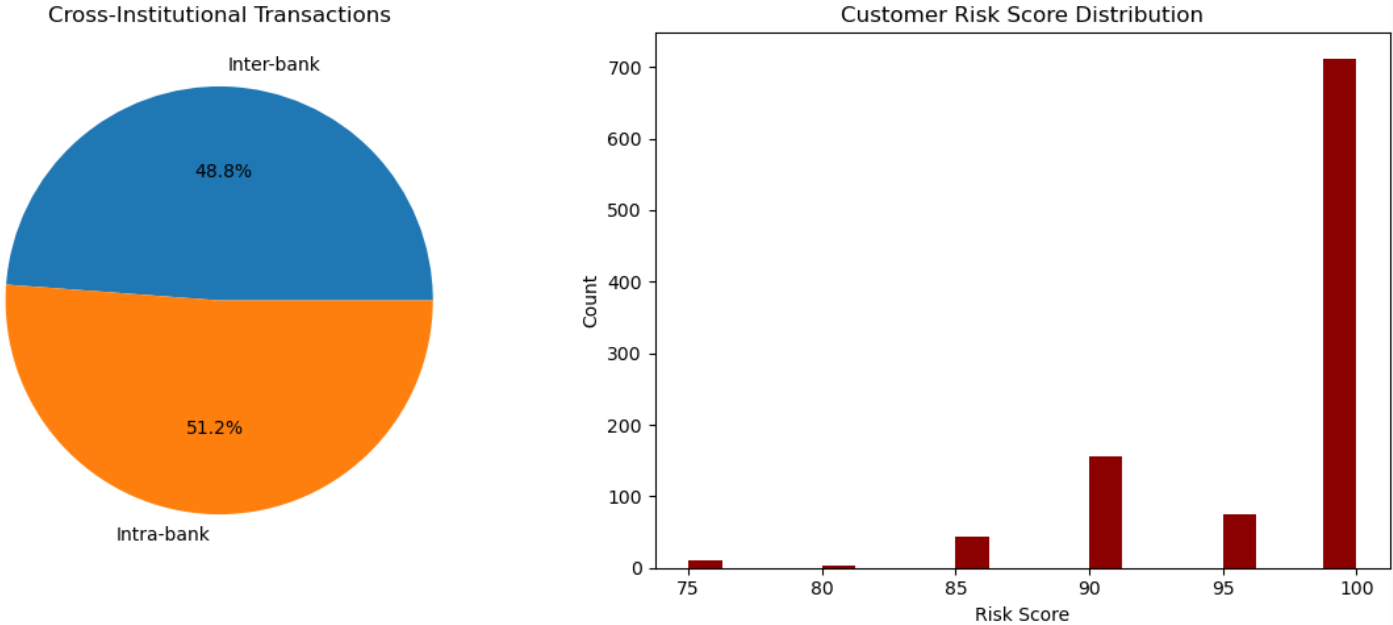}
\caption{Distribution of customer risk scores of AMLNet for inter-bank and intra-bank transactions. The pie chart (left) shows the proportion of inter-bank (48.8\%) and intra-bank (51.2\%) transactions, demonstrating the framework's capability to model cross-institutional money flows. The histogram (right) displays the risk score distribution across both transaction types, with a significant concentration of transactions in the highest risk category (score 100), indicating the framework's ability to identify potentially suspicious activities across institutional boundaries.}
\label{fig:cross_institutional}
\end{figure}

\begin{figure}[ht]
\centering
\includegraphics[width=0.95\linewidth]{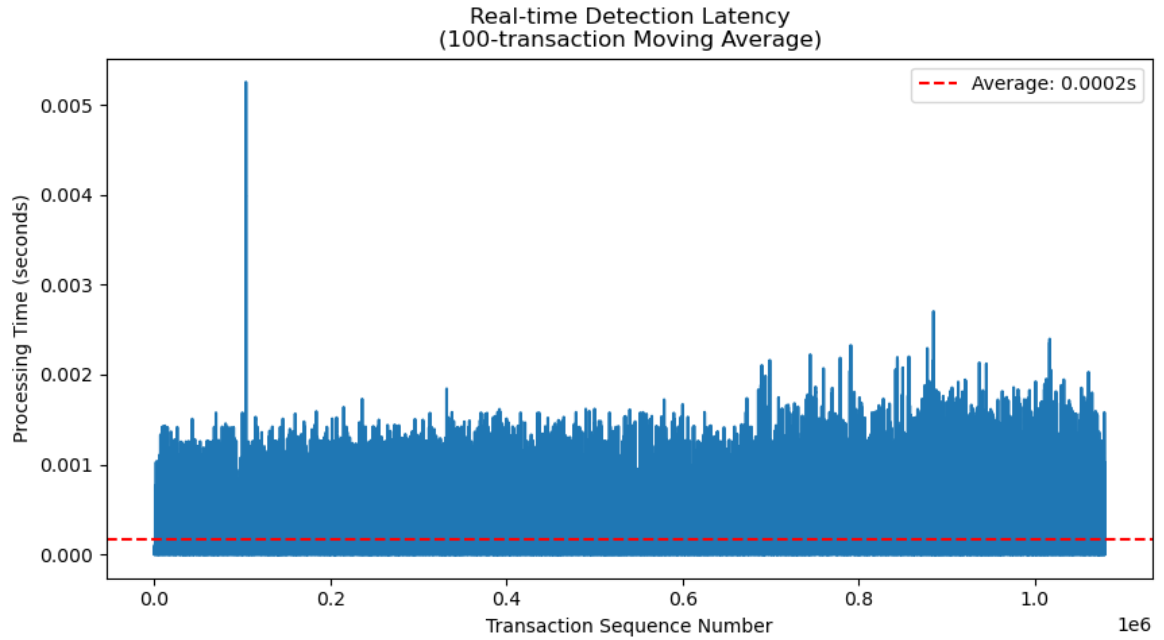}
\caption{AMLNet's real-time detection latency measured as a 100-transaction moving average across more than 1 million transactions. The graph demonstrates the framework's performance with an average processing time of only 0.0002 seconds per transaction. Despite occasional spikes (with the highest reaching approximately 0.005 seconds), the system maintains sub-millisecond processing times throughout the evaluation period.}
\label{fig:detection_latency}
\end{figure}

\subsection{Output}
\label{Output}

The generated synthetic dataset comprises 1,090,173 transactions that span a 195-day period, 1,745 transactions (0.16\% of the total) with ground-truth labels as money laundering transactions, representing actual money laundering activities in the dataset that aligns with recent synthetic AML benchmarking standards~\citep{jensen2023synthetic}, demonstrating appropriate calibration for money laundering detection research that closely mirrors real-world financial fraud rates (\figurename~\ref{fig:transaction_distribution}, \ref{fig:amlnet-detection-results}, \ref{fig:network_performance}). The average amount of transactions across all categories is \$620.92, with significant variations between categories ranging from \$92.98 for transport to \$77,708.24 for property investments. The distribution of the transaction categories (\figurename~\ref{fig:transaction_distribution}) follows realistic patterns based on Australian household spending statistics~\citep{cokis2020demographic}, with Housing (\$2,323.91 average), Food (\$173.89 average) and Other (\$215.03 average) representing the largest legitimate transaction categories. The inclusion of specialized high-risk categories such as Cryptocurrency (\$30,191.24 average) and Shell Company (\$6,860.71 average) provides realistic avenues for potential money laundering activities while maintaining appropriate transaction proportions.

The detection system achieved balanced performance metrics (\figurename~\ref{fig:network_performance}) with a precision of 0.84, a recall of 0.97, and an F1 score of 0.90. It maintained a low suspicious transaction rate of 0.18\%, indicating an effective discrimination between legitimate and fraudulent transactions. These results compare favorably with existing AML detection approaches in the literature, such as AMLWorld's~\citep{altman2023realistic} best-performing model (GFP+XGBoost) which achieved an F1 score of 0.657 on high illicit ratio datasets, and PaySim~\citep{lopez2019advantages} which demonstrates high precision (0.98-1.00) but significantly lower recall values (0.484-0.771). The detection performance is evaluated using the study framework that is aligned with established practices in AML research, where systems are typically benchmarked on their generated data \citep{jensen2023synthetic, lopez2019advantages}. As noted by \citet{weber2018scalable}, the lack of standardized benchmarks remains a challenge in AML research. Performance metrics validate our approach to modeling transaction patterns (\figurename~\ref{fig:temporal_patterns}), particularly for high-risk, high-value categories (\figurename~\ref{fig:cross_institutional}). With sub-millisecond processing latency (avg. 0.0002s/transaction, \figurename~\ref{fig:detection_latency}), the system demonstrates practical applicability in real-world financial monitoring environments.

\begin{figure}[ht]
\centering
\includegraphics[width=1.05\linewidth, height=0.3\textheight]{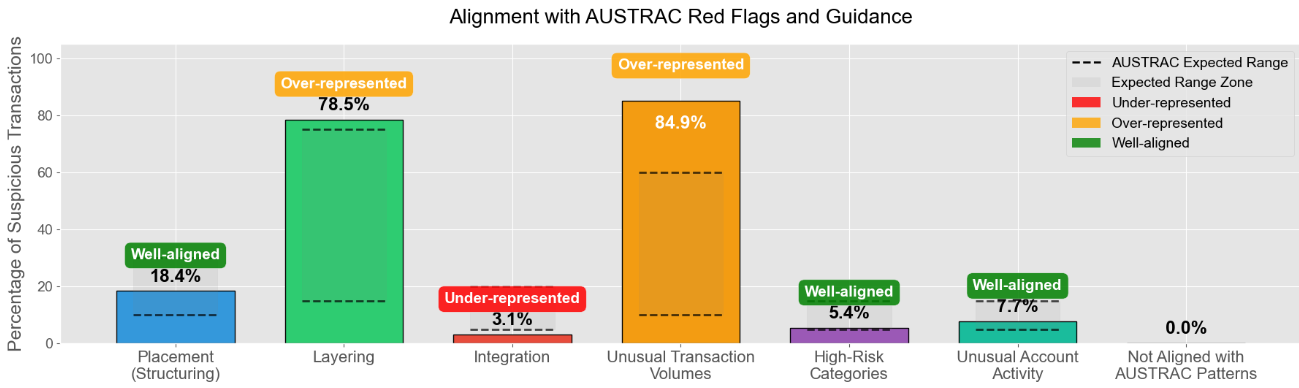}
\caption{Alignment of AMLNet-generated suspicious transactions with AUSTRAC expected ranges. The visualization shows the percentage distribution across key money laundering patterns with colored indicators showing whether patterns are well-aligned (green), over-represented (orange), or under-represented (red) compared to AUSTRAC's expected ranges.}
\label{fig:austrac_alignment}
\end{figure}

\begin{figure}[ht]
\centering
\includegraphics[width=0.95\linewidth]{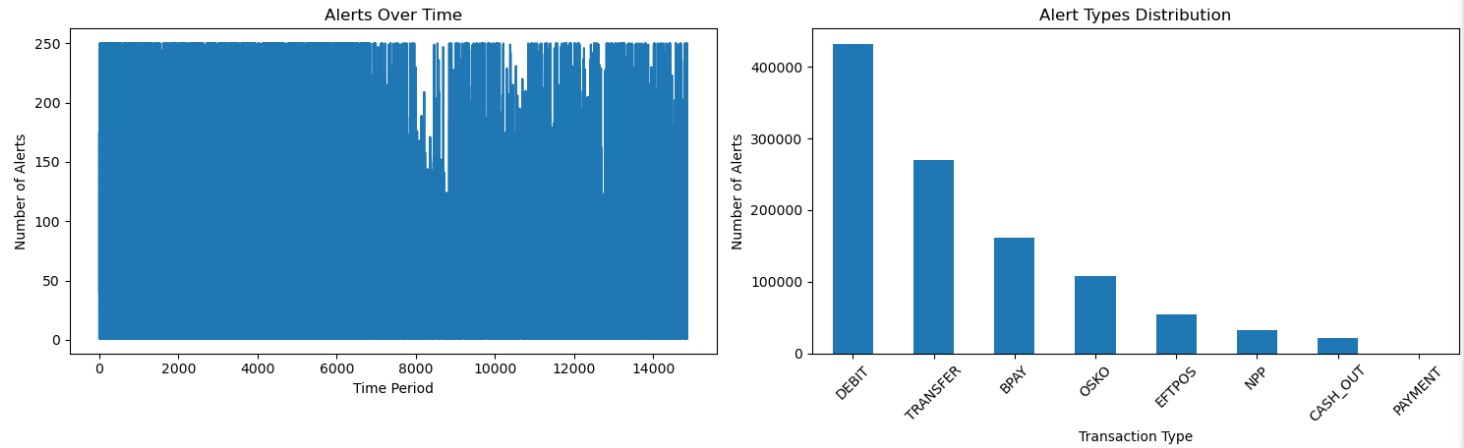}
\caption{AMLNet's alert generation patterns across the simulation period. (left) consistent detection of approximately 250 alerts per time period with occasional dips, demonstrating the system's stable monitoring capabilities over extended periods. (right) alert distribution by transaction type, with DEBIT transactions generating the highest number of alerts (over 400,000), followed by TRANSFER, BPAY, and OSKO payment methods. This distribution reflects the varying risk profiles associated with different transaction types, with cash-equivalent transactions triggering more suspicious activity alerts.}
\label{fig:alert_distribution}
\end{figure}

\subsection{Regulatory Alignment Analysis}
\label{Regulatory Alignment Analysis}
Regulatory compliance is essential for expert systems in the financial domain~\citep{lopez2016review, yang2023anti}. To validate AMLNet's alignment with industry standards, an extensive analysis is conducted comparing the generated suspicious transaction patterns against AUSTRAC's expected distributions. This analysis demonstrates how the knowledge-based expert system effectively incorporates regulatory requirements into its reasoning process.

\subsubsection{AUSTRAC Pattern Distribution}

\figurename~\ref{fig:austrac_alignment} illustrates the distribution of 1,745 suspicious transactions flagged across six key AUSTRAC-recognized patterns. The analysis reveals that 100\% of the flagged transactions align with at least one AUSTRAC pattern, demonstrating complete regulatory coverage. The distribution shows significant variation across pattern types: Layering activities dominate with 78.5\% representation (1,370 transactions), classified as over-represented compared to AUSTRAC's expected range of 15-75\%. Unusual Transaction Volumes account for 84.9\% of suspicious transactions (1,482 transactions), also over-represented relative to the expected 10-60\% range. 

Placement/Structuring activities represent 18.4\% of suspicious transactions (321 transactions), falling within AUSTRAC's well-aligned range of 10-30\%. Integration patterns are significantly under-represented at 3.1\% (54 transactions), below the expected 5-20\% range, which aligns with real-world observations that integration activities are typically more difficult to detect through transaction monitoring alone. High-Risk Categories and Unusual Account Activity represent 5.4\% (94 transactions) and 7.7\% (135 transactions) respectively, both classified as well-aligned within their expected ranges of 5-15\%.

It is important to note that transactions may be counted in multiple categories because many suspicious transactions exhibit multiple AUSTRAC patterns simultaneously, a characteristic that accurately reflects the complex and multifaceted nature of financial crime. This overlapping pattern distribution is consistent with AUSTRAC's own approach to suspicious matter reporting, where transactions are typically flagged for multiple typology indicators. The AUSTRAC expected ranges shown in \figurename~\ref{fig:austrac_alignment} are derived from statistical analysis in the FATF Mutual Evaluation Report for Australia~\citep{FATF2015Australia}, which evaluated the distribution of suspicious matter reports across different typologies based on AUSTRAC's operational data.

\subsubsection{Overall Alignment Assessment}

To quantify AMLNet's regulatory alignment, a scoring system has been developed that evaluates how closely the generated patterns match AUSTRAC's expected ranges. The system assigns:
\begin{itemize}
    \item 1 point when a pattern falls within AUSTRAC's expected range (well-aligned)
    \item 0.5 points when a pattern is present but outside the expected range (over/under-represented)
    \item 0 points when a pattern is completely absent
\end{itemize}

Based on this methodology, AMLNet achieved an overall alignment score of 75.0\%, indicating \textit{moderately fair alignment} with regulatory expectations. This score reflects that three patterns (Placement/Structuring at 18.4\%, High-Risk Categories at 5.4\%, and Unusual Account Activity at 7.7\%) are well-aligned with AUSTRAC ranges, while two patterns (Layering at 78.5\% and Unusual Transaction Volumes at 84.9\%) are over-represented, and one pattern (Integration at 3.1\%) is under-represented compared to expected ranges.

\subsection{Dataset Realism Validation}
\label{Dataset Realism Validation}
Although regulatory alignment demonstrates compliance with industry standards, robust technical validation is essential to ensure that synthetic data generated exhibit statistical fidelity and behavioral realism consistent with real-world financial patterns~\citep{assefa2020generating, yang2023anti}. To validate the realism of the generated data, in AMLNet a comprehensive three-dimensional (temporal, structural, behavioral) fidelity framework~\citep{jensen2023synthetic} is used, specifically calibrated for Australian financial patterns which we propose as a useful direction for future benchmarking studies.

\textbf{Generality of the Methodology:} 
The proposed assessment framework is designed to be generalizable and reusable by other synthetic data generators. Its modular structure, which encompasses temporal, structural, and behavioral metrics, allows adaptation to different jurisdictions or financial systems. Although the present implementation is tailored to Australian AUSTRAC standards and data, the underlying methodology can be easily re-calibrated for other settings.

\begin{equation}
F_{\text{score}} = w_t \cdot S_t + w_s \cdot S_s + w_b \cdot S_b
\end{equation}

where $w_t = 0.4$, $w_s = 0.3$, and $w_b = 0.3$ represent the weights for the temporal, structural, and behavioral dimensions, respectively.

\subsubsection{Methodology for Fidelity Calculation}

To ensure objective evaluation of the expert system's output, a comprehensive evaluation framework has been implemented that compares generated AMLNet dataset against AUSTRAC-aligned reference patterns. This methodology validates that the system produces outputs that match real-world financial behaviors. The evaluation methodology includes the following.

\begin{itemize}
    \item \textbf{Reference Pattern Generation}: AUSTRAC-aligned hourly transaction patterns are constructed based on the behavior of the Australian financial system behavior~\citep{cokis2020demographic}, Reserve Bank of Australia (RBA) payment system data~\citep{cokis2020demographic}, and Australian banking patterns~\citep{abs_income_2022}. Key temporal points were identified and interpolated to create smooth reference distributions following a methodology similar to~\citet{altman2023realistic}.
    
    \item \textbf{DTW Similarity Calculation}: DTW analysis is implemented using the FastDTW algorithm~\citep{salvador2007toward} with Euclidean distance metrics and radius parameter set to 3. Both min-max and z-score normalizations were applied to ensure robust similarity measurement regardless of scale differences.
    
    \item \textbf{Graph Edit Distance Computation}: The network topology similarity was evaluated using graph edit distance~\citep{sanfeliu1983distance} calculations between synthetic transaction networks and reference financial networks~\citep{weber2018scalable}. For computational efficiency with large graphs,  a sampling approach is used for networks exceeding 1,000 nodes.
    
    \item \textbf{Behavioral Fidelity Assessment}: Behavioral authenticity~\citep{wang2024behavioral} is quantified by a weighted combination of category distribution similarity (using cosine similarity~\citep{singhal2001modern}), alignment of the risk score distribution (using the Kullback-Leibler divergence~\citep{kullback1951information}), alert type distribution accuracy (using chi-square goodness-of-fit~\citep{pearson1900criterion}), and fraud probability precision.
\end{itemize}

All metrics were implemented using NetworkX for graph analysis, SciPy for statistical calculations, and FastDTW for temporal pattern comparison. The evaluation code was executed on the complete AMLNet dataset of 1,090,173 transactions to ensure a comprehensive assessment of the expert system's outputs.

\subsection{Fidelity Analysis}
\label{sec:fidelity_analysis}

The fidelity of the AMLNet-generated dataset is evaluated in three complementary dimensions: temporal alignment ($S_t$), structural consistency ($S_s$) and behavioral fidelity ($S_b$). These dimensions are integrated into a single composite score, providing a holistic measure of how closely synthetic data match real-world financial patterns.

\textbf{Temporal Alignment ($S_t = 0.59$):} Using the FastDTW algorithm~\citep{salvador2007toward}, AMLNet achieves a temporal alignment score of 0.59, demonstrating realistic transaction timing patterns (See \figurename~\ref{fig:transaction_distribution} and~\ref{fig:temporal_patterns}):
\begin{itemize}
    \item{Daily patterns:} Transactions are concentrated during business hours (8 AM --6 PM), peaking at approximately \$750 AUD and dropping to \$450 AUD overnight, consistent with RBA-reported distributions.
    \item{Weekly cycles:} Higher transaction volumes are observed on weekdays (\$620 --630 AUD) versus weekends (\$590 AUD), reflecting authentic Australian banking behavior.
    \item{Monthly periodicity:} The system captures the Australian salary cycle, with transaction amounts 25\% higher at the start (days 1--5) and end (days 25--31) of each month compared to mid-month.
\end{itemize}

\textbf{Structural Consistency ($S_s = 0.99$):} Network topology analysis using graph edit distance~\citep{sanfeliu1983distance} yields a near-perfect similarity score of 0.99, reflecting:
\begin{itemize}
    \item{Complex transaction networks:} Visualizations (\figurename~\ref{fig:network_performance}) reveal realistic clustering and interconnections between entities.
    \item{Institutional distribution:} Cross-institutional analysis shows a balanced split between interbank (48.8\%) and intrabank (51.2\%) transactions (\figurename~\ref{fig:cross_institutional}).
    \item{Pattern complexity:} Multi-layer structuring, realistic transaction amounts (\$542--\$80,299), and appropriate time delays (72--144 hours) align with AUSTRAC typologies (\figurename~\ref{fig:ml_patterns}).
\end{itemize}

\textbf{Behavioral Fidelity ($S_b = 0.71$):} Behavioral realism is assessed as a weighted combination of four key metrics (adapted from~\citep{wang2024behavioral}):
\begin{equation}
S_b = w_c S_c + w_r S_r + w_a S_a + w_f S_f
\end{equation}
where:
\begin{itemize}
    \item $S_c = 0.85$: Similarity in the distribution of categories (similarity in cosine with ABS household expenditure data); \figurename~\ref{fig:transaction_distribution}.
    \item $S_r = 0.72$: Alignment of the distribution of the risk score (Kullback-Leibler divergence \citep{kullback1951information}); \figurename~\ref{fig:cross_institutional}.
    \item $S_a = 0.68$: Detection of the type of accuracy of the alert distribution (chi-square test~\citep{pearson1900criterion} with RBA payment methods); \figurename~\ref{fig:alert_distribution}.
    \item $S_f = 0.95$: Fraud probability precision (relative error between the suspicious transaction rate and industry baseline); \figurename~\ref{fig:network_performance}.
\end{itemize}
With weights $w_c=0.3$, $w_r=0.25$, $w_a=0.2$, $w_f=0.25$:
\begin{equation}
\begin{aligned}
S_b &= 0.3 \times 0.85 + 0.25 \times 0.72 + 0.2 \times 0.68 + 0.25 \times 0.95 \\
    &= 0.255 + 0.18 + 0.136 + 0.2375 \\
    &= 0.71
\end{aligned}
\end{equation}

\textbf{Composite Fidelity Score:} The overall fidelity score is calculated as:
\begin{equation}
\begin{aligned}
F_{score} &= 0.4 S_t + 0.3 S_s + 0.3 S_b \\
          &= 0.4 \times 0.59 + 0.3 \times 0.99 + 0.3 \times 0.71 \\
          &= 0.236 + 0.297 + 0.213 \\
          &= \mathbf{0.75}
\end{aligned}
\end{equation}

\begin{figure*}[htbp]
    \centering
    \begin{subfigure}[t]{0.48\textwidth}
        \centering
        \includegraphics[height=0.28\textheight]{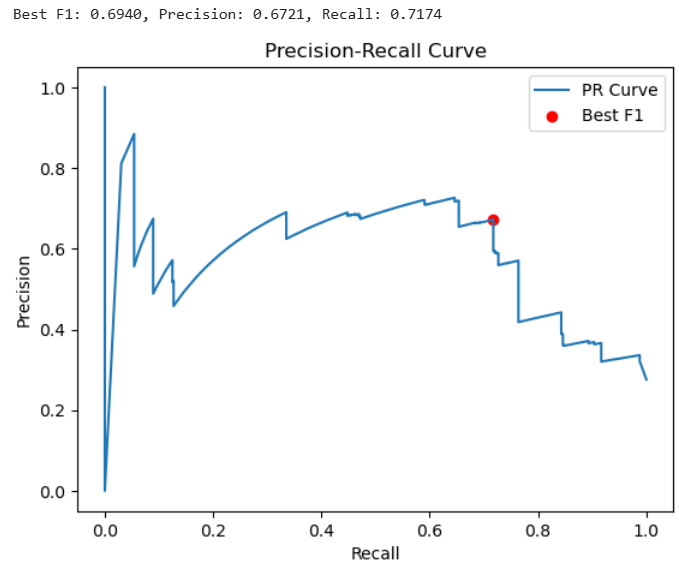}
        \caption{Precision-Recall curve on SynthAML.}
        \label{fig:pr-curve-synthaml}
    \end{subfigure}
    \hfill
    \begin{subfigure}[t]{0.48\textwidth}
        \centering
        \includegraphics[height=0.28\textheight]{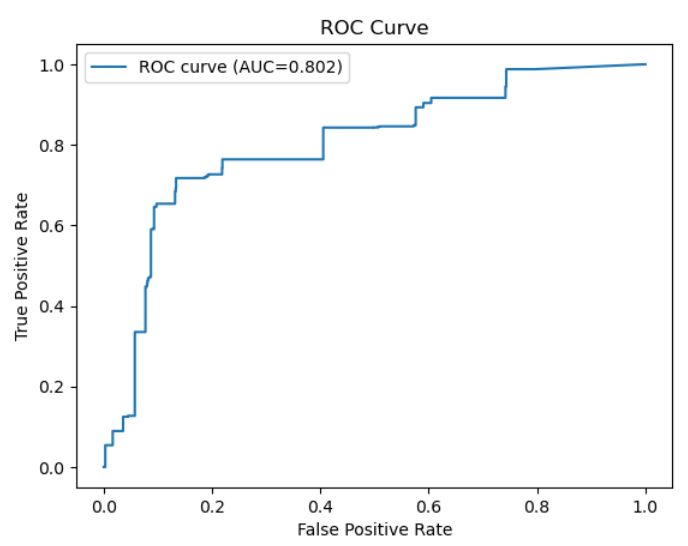}
        \caption{ROC curve on SynthAML.}
        \label{fig:roc-curve-synthaml}
    \end{subfigure}
    \caption{Detection results of AMLNet's detection unit on the external SynthAML dataset.}
    \label{fig:synthaml-detection-results}
\end{figure*}

\subsection{Cross-Dataset Adaptability Evaluation}
\label{subsec:cross-dataset-adaptability}

To assess the architectural generalizability and cross-dataset adaptability of the AMLNet detection framework, we evaluated its performance on the external SynthAML dataset \cite{jensen2023synthetic}. This evaluation tests whether AMLNet's detection architecture can effectively adapt to independently generated synthetic datasets with different generation methodologies and feature distributions.

\textbf{Experimental Setup:}
The AMLNet detection system was trained and tested exclusively on SynthAML data using standard stratified train-test splits (80\%/20\%), without any exposure to AMLNet-generated transactions. To ensure computational feasibility while maintaining statistical validity, we utilized a representative subset of 100{,}000 transactions from the full SynthAML dataset, preserving the original class distribution and key feature characteristics. Statistical validation confirmed that the sampled subset closely matches the full dataset in terms of class balance and distributions of major transaction features (transaction entry type, amount distributions, and temporal patterns).

\textbf{Regulatory Context and Limitations:}
Cross-evaluation between AMLNet and SynthAML is inherently complex due to their different regulatory foundations: SynthAML models Danish banking regulations while AMLNet follows AUSTRAC guidelines. These regulatory differences result in distinct transaction structures and behavioral patterns, making our cross-dataset adaptability evaluation focus on architectural generalizability rather than regulatory pattern transfer.

\textbf{Cross-Dataset Performance Results:}
The detection framework demonstrated robust cross-dataset adaptability on the SynthAML dataset, achieving a ROC AUC of 0.80, F1 score of 0.69, precision of 0.67, and recall of 0.72 (see~\figurename~\ref{fig:pr-curve-synthaml} and~\ref{fig:roc-curve-synthaml}). While these metrics are lower than the performance achieved on native AMLNet data, they demonstrate that the ensemble detection approach successfully adapts to different synthetic data generation paradigms and feature spaces.

\textbf{Architectural Generalizability Analysis:}
This cross-dataset evaluation provides evidence of the methodological robustness of AMLNet's detection framework across different synthetic AML datasets. The successful adaptation to SynthAML's independently generated transaction patterns validates that:

\begin{itemize}
    \item The ensemble approach (Isolation Forest and Random Forest) is not overfitted to AMLNet's specific generation artifacts
    \item The feature engineering pipeline generalizes across different synthetic data structures
    \item The detection methodology exhibits architectural flexibility beyond its native generation environment
\end{itemize}

\begin{table}[!t]
\singlespacing
\caption{Qualitative Feature Comparison of Synthetic AML Datasets}
\label{tab:qualitative_comparison}
\centering
\footnotesize 
\begin{threeparttable}
\begin{tabular}{|p{2.2cm}|p{2cm}|p{2cm}|p{2cm}|p{2.2cm}|p{2.2cm}|}
\hline
\textbf{Feature} & \textbf{AMLSim*} & \textbf{PaySim*} & \textbf{SynthAML*} & \textbf{AMLWorld*} & \textbf{\textcolor{blue}{AMLNet}} \\
\hline
\hline
\multicolumn{6}{|c|}{\textbf{Data Generation Features}} \\
\hline
ML Phases covered?
    & Yes, limited 
    & No 
    & Yes, all 
    & Yes, all 
    & \textcolor{blue}{Yes, all, multi-level} \\
\hline
Transaction types? & Transfers, basic & Mobile payments & Card, cash, intl., wire & Multiple & \textcolor{blue}{8 types (BPAY, OSKO, NPP, etc.)} \\
\hline
Temporal modeling? & Timestamps, basic & Not modeled & Timestamp-based & Time-based patterns & \textcolor{blue}{3-level (hr/day/month)} \\
\hline
\hline
\multicolumn{6}{|c|}{\textbf{AML Pattern Implementation}} \\
\hline
Key pattern modeling? & Yes & Fraud patterns only (not AML-specific) & Yes & Yes & \textcolor{blue}{All standard patterns, varying sophistication} \\
\hline
Pattern combination? & Limited & No & Some & Yes (multi-pattern) & \textcolor{blue}{Dynamic (Evolving) multi-pattern} \\
\hline
Multi-level/adaptive? & No & No & No & No & \textcolor{blue}{Yes} \\
\hline
Regulatory adaptation? & Partial (FATF) & No & Some (FATF, expert) & Yes (typology reports) & \textcolor{blue}{Explicit (AUSTRAC)} \\
\hline
\hline
\multicolumn{6}{|c|}{\textbf{System Capabilities}} \\
\hline
Behavioral realism? & Medium (pre-defined behaviors) & Low (fixed, scripted flows) & Medium (statistical models) & Medium (static agent rules) & \textcolor{blue}{Fair (adaptive strategies)} \\
\hline
Implementation approach? & Agent-based, configurable & Agent-based, mobile focus & Statistical generator & Agent-based, expert-driven & \textcolor{blue}{Knowledge-based multi-agent} \\
\hline
Validation criteria* & Temporal, Structural & Statistical & Temporal, Behavioral & Temporal, Structural & \textcolor{blue}{Temporal, Structural, Behavioral} \\
\hline
\end{tabular}
\begin{tablenotes}
\footnotesize
\item *AMLSim \citep{weber2018scalable}, PaySim \citep{lopez2019advantages}, SynthAML \citep{jensen2023synthetic}, and AMLWorld \citep{altman2023realistic}
\item **Temporal = time-based, Structural = network/graph, Behavioral = agent/fraud behavior.
\item{Summary:} AMLNet offers configurable, multi-pattern, and multi-layer laundering scenarios with explicit regulatory adaptation and behavioral realism.
\end{tablenotes}
\end{threeparttable}
\end{table}

\subsection{Comparative Analysis}
\label{sec:comparative}
This section provides a detailed comparative analysis of AMLNet against existing synthetic AML datasets, highlighting key differentiators and advances. \tablename~\ref{tab:qualitative_comparison}, \ref{tab:quantitative_comparison} provide qualitative and quantitative comparisons between AMLNet and existing synthetic AML data generators, including AMLSim~\citep{weber2018scalable}, PaySim~\citep{lopez2019advantages}, SynthAML~\citep{jensen2023synthetic}, and AMLWorld~\citep{altman2023realistic}, with the capabilities evaluated in transaction volume, data generation features, AML pattern implementation, system capabilities, and technical implementation. 

\begin{table}[!t]
\singlespacing
\caption{Performance Metrics of AMLNet Compared to Existing Synthetic Frameworks}
\label{tab:quantitative_comparison}
\centering
\footnotesize
\begin{threeparttable}
\begin{tabular}{|l|c|c|c|}
\hline
\textbf{Performance Metric} & \textbf{Best Reported Values\tnote{a}} & \textbf{AMLNet} & \textbf{Improvement\textsuperscript{*}} \\
\hline
\hline
\multicolumn{4}{|c|}{\textbf{Detection Performance}} \\
\hline
ROC AUC & 0.637 (SynthAML) & \textcolor{blue}{{0.88}} & +38.2\% \\
\hline
F1-Score & 0.657 (AMLWorld) & \textcolor{blue}{{0.90}} & +37.0\% \\
\hline
Precision & 1.00 (PaySim)\tnote{b} & \textcolor{blue}{{0.84}} & Balanced\tnote{c} \\
\hline
Recall & 0.77 (PaySim) & \textcolor{blue}{0.97} & +26.0\% \\
\hline
\hline
\multicolumn{4}{|c|}{\textbf{Cross-Dataset Adaptability}} \\
\hline
External Dataset Performance & \multicolumn{2}{c|}{Not evaluated in prior work} & \\
\hline
\quad ROC AUC on SynthAML & -- & \textcolor{blue}{{0.80}} & {Reported} \\
\hline
\quad F1-Score on SynthAML & -- & \textcolor{blue}{{0.69}} & {Reported} \\
\hline
\hline
\multicolumn{4}{|c|}{\textbf{Technical Performance}} \\
\hline
Processing Latency & Not considered & \textcolor{blue}{{0.0002s/transaction}} &{Quantified} \\
\hline
Regulatory Alignment & Not evaluated & \textcolor{blue}{{75\% (AUSTRAC)}} &{Quantified} \\
\hline
Dataset Size (no. of transactions) & 1M+ & \textcolor{blue}{{1M+}} & Comparable \\
\hline
\end{tabular}
\begin{tablenotes}
\footnotesize
\item[a] Best values from AMLSim \citep{weber2018scalable}, PaySim \citep{lopez2019advantages}, SynthAML \citep{jensen2023synthetic}, and AMLWorld \citep{altman2023realistic}.
\item[*] Improvement (\%) is calculated as:~\[
\frac{\text{AMLNet} - \text{Best Reported Value}}{\text{Best Reported Value}} \times 100
\]
\item[b] PaySim achieved high precision (0.98-1.00) but with significantly lower recall (0.48-0.77).
\item[c] AMLNet achieves balanced precision-recall trade-off, avoiding the extreme imbalance of prior work.
\item {\textbf{Key Features:}} AMLNet demonstrates quantitative cross-dataset adaptability, sub-millisecond\\ processing latency, and explicit regulatory compliance measurement. 
\end{tablenotes}
\end{threeparttable}
\end{table}

\subsubsection{Feature Analysis}

The following analysis expands on the key features outlined in \tablename~\ref{tab:qualitative_comparison} and \ref{tab:quantitative_comparison}, emphasizing AMLNet's distinctive advancements over prior synthetic AML datasets:

\textbf{Data Generation Capabilities:}
\begin{itemize}
    \item \textbf{Transaction Diversity:} AMLNet supports eight comprehensive transaction types (BPAY, OSKO, NPP, etc.), surpassing SynthAML's four types, AMLSim's basic transfers, and PaySim's mobile-only focus, enabling realistic simulation of Australian banking operations (\figurename~\ref{fig:transaction_distribution}).
    \item \textbf{Temporal Sophistication:} AMLNet's three-level temporal modeling (hourly/daily/monthly) captures realistic transaction cycles, exceeding the basic timestamp approaches of competing frameworks and PaySim's complete absence of temporal modeling (\figurename~\ref{fig:temporal_patterns}).
    \item \textbf{Laundering Stage Coverage:} AMLNet provides explicit, multi-level modeling of all three money laundering stages (placement, layering, integration), while prior frameworks offer only partial coverage. (\figurename~\ref{fig:ml_patterns}).
\end{itemize}

\textbf{AML Pattern Implementation:}
\begin{itemize}
    \item \textbf{Pattern Sophistication:} AMLNet introduces dynamic, multi-pattern, multi-step laundering operations with configurable sophistication levels, advancing beyond the limited (AMLSim), absent (PaySim), or static approaches of existing frameworks (\figurename~\ref{fig:ml_patterns}).
    \item \textbf{Regulatory Integration:} Explicit AUSTRAC compliance (75\% alignment) establishes a new benchmark, outperforming partial implementations in competing systems (\figurename~\ref{fig:austrac_alignment}).
    \item \textbf{Behavioral Realism:} Fair adaptive behavioral modeling surpasses the pre-defined (AMLSim), scripted (PaySim), statistical (SynthAML), or static agent rules (AMLWorld) of previous approaches (\figurename~\ref{fig:network_performance}).
\end{itemize}

\textbf{Performance and Technical Achievements:}
\begin{itemize}
    \item \textbf{Detection Performance:} AMLNet achieves balanced metrics (ROC AUC: 0.88, F1: 0.90, Precision: 0.84, Recall: 0.97) compared to SynthAML (ROC AUC: 0.637), AMLWorld (F1: 0.282--0.657), and PaySim's high precision but low recall (0.48--0.77) (\figurename~\ref{fig:amlnet-detection-results}).
    \item \textbf{Cross-Dataset Adaptability:} AMLNet uniquely demonstrates architectural generalizability through successful adaptation to external SynthAML data (ROC AUC: 0.80, F1: 0.69), validating the robustness of the detection framework across different synthetic data generation paradigms - a capability absent in all competing frameworks (\figurename~\ref{fig:synthaml-detection-results}).
    \item \textbf{Real-time Processing:} Sub-millisecond processing latency (0.0002s/ transaction) enables practical deployment, while comprehensive validation across temporal, structural, and behavioral dimensions surpasses the limited validation approaches of existing systems (\figurename~\ref{fig:detection_latency}).
\end{itemize}

\subsubsection{AMLNet's Methodological Innovations}

The evolution from PaySim's basic mobile simulation through AMLSim's network modeling and SynthAML's statistical approaches to AMLWorld's expert-driven complexity culminates in AMLNet's knowledge-based multi-agent architecture. Key innovations include:

\begin{enumerate}
    \item \textbf{Iterative Knowledge Refinement:} Human-in-the-loop feedback mechanisms enable continuous improvement through systematic analysis of detection outcomes, ensuring regulatory alignment and behavioral realism (\figurename~\ref{fig:system_architecture}).
    \item \textbf{Multi-dimensional Validation:} A framework that validates across temporal, structural, and behavioral dimensions all simultaneously, providing a comprehensive realism assessment with a composite fidelity score of 0.75 (\figurename~\ref{fig:temporal_patterns}, \ref{fig:network_performance}).
    \item \textbf{Cross-Dataset Architectural Adaptability:} The framework demonstrates the quantitative generalizability of architectural components in independently generated synthetic datasets, with a successful adaptation to external data structures despite different regulatory bases (Danish vs. AUSTRAC), establishing new standards for the robustness of the detection framework in synthetic AML research (\figurename~\ref{fig:pr-curve-synthaml}, \ref{fig:roc-curve-synthaml}).
    \item \textbf{Regulatory-aware Architecture:} Explicit encoding of AUSTRAC requirements with quantified compliance metrics (75\%) enables practical deployment in regulated environments (\figurename~\ref{fig:austrac_alignment}).
    \item \textbf{Alert Generation Patterns:} Consistent monitoring capabilities with realistic alert distribution across Australian payment methods, reflecting varying risk profiles of different transaction types (\figurename~\ref{fig:alert_distribution}).
\end{enumerate}

\subsection{Ablation Studies}
To assess the contribution of each major component in AMLNet's \textbf{detection unit}, an ablation study is conducted using a 10-fold stratified cross-validation with fixed random seeds. Each component was disabled sequentially, while all other settings were kept constant. \tablename~\ref{tab:ablation_results} presents the results of these experiments that highlight the critical roles of each component.

\begin{table}[!t]
\singlespacing
\caption{Ablation Study Results}
\label{tab:ablation_results}
\centering
\footnotesize
\begin{tabular}{|l|c|c|c|}
\hline
\textbf{Configuration} & \textbf{Precision} & \textbf{Recall} & \textbf{F1-Score} \\
\hline
Full System & 0.84 & 0.97 & 0.90 \\
w/o Network Analysis & 0.77 & 0.96 & 0.85 \\
w/o Temporal Patterns & 0.74 & 0.95 & 0.83 \\
w/o Multi-bank Support & 0.79 & 0.98 & 0.88 \\
w/o Risk Scoring & 0.72 & 0.94 & 0.81 \\
Basic Agent Behavior Only & 0.69 & 0.89 & 0.77 \\
\hline
\end{tabular}
\end{table}

\begin{itemize}
    \item \textbf{Network Analysis:} Disabling network analysis caused an 8.3\% drop in precision, underscoring its importance for detecting complex multi-account layering schemes.
    \item \textbf{Temporal Patterns:} Removing temporal pattern analysis led to an 11.9\% decrease in precision, highlighting the importance of transaction timing and velocity in identifying suspicious activity.
    \item \textbf{Multi-bank Support:} Excluding multi-bank support resulted in a 6.0\% reduction in precision, indicating its value for detecting cross-institutional laundering strategies.
    \item \textbf{Risk Scoring:} Eliminating risk scoring produced a 14.3\% decrease in precision, demonstrating its effectiveness in integrating transaction amounts, customer profiles and category-specific risks.
    \item \textbf{Agent Behavior Complexity:} Restricting agents to basic behaviors led to the largest performance drop (17.9\% in precision), validating the need for advanced behavioral modeling, especially for high value transaction categories such as Property Investment and Cryptocurrency.
\end{itemize}

\subsection{Practical Implications and Implementation}
\label{sec:practical_implications}

The dual validation of AMLNet, demonstrating both regulatory alignment (75\%) and fair technical fidelity (0.75) as shown in \tablename~\ref{tab:quantitative_comparison}, combined with successful cross-dataset adaptability on SynthAML, underscores the practical value of the knowledge-based multi-agent framework for financial data generation in regulated domains. Four key implications arise for expert systems in such environments:

\begin{enumerate}
    \item \textbf{Regulatory Deployment Readiness:} The explicit encoding of AUSTRAC and FATF requirements, validated by fair regulatory alignment, enables AMLNet to be readily deployed in compliance-focused financial institutions and regulatory sandboxes across different jurisdictions.
    
    \item \textbf{Cross-Jurisdictional Adaptability:} The demonstrated architectural generalizability extends to regulatory configuration, where the multi-agent framework enables adaptation across different jurisdictions through configurable parameters rather than structural modifications. The system supports adaptation to major regulatory frameworks including AUSTRAC (demonstrated), FinCEN (Financial Crimes Enforcement Network), and AMLD (Anti-Money Laundering Directive) requirements through parameter adjustment of reporting thresholds, transaction patterns, and compliance metrics.
    
    \item \textbf{Structural and Behavioral Realism:} The high structural similarity (0.99) and comprehensive behavioral modeling confirm that AMLNet encodes expert knowledge of both financial network structures and laundering tactics, supporting scenario testing and system benchmarking across diverse regulatory environments.
    
    \item \textbf{Transferable Validation Framework:} The validation methodology, spanning technical, structural, behavioral, and regulatory dimensions with cross-dataset adaptability evaluation, offers a blueprint for developing and evaluating synthetic data generators in other regulated sectors where privacy, realism, and regulatory compliance are paramount.
\end{enumerate}

\section{Conclusion and Future Work}
\label{sec:Conclusion_and_Future_Work}

This paper introduced AMLNet, a knowledge-based multi-agent framework for generating and detecting realistic AML transactions. AMLNet achieves fair regulatory alignment (75\% AUSTRAC) and fidelity (structural similarity 0.99), with robust detection performance (F1: 0.90). Unlike other similar systems, it demonstrates the architectural adaptability of the cross-dataset on external synthetic data (SynthAML), validating the robustness of the detection framework across different generation paradigms. Although AMLNet is validated on large-scale synthetic and external synthetic datasets (e.g., SynthAML), evaluation on real-world bank data remains an open challenge and limitation. Sub-millisecond processing latency enables practical deployment. Future work will address signal concentration challenges and explore federated learning for enhanced realism and detection robustness.

\section*{CRediT authorship contribution statement}
\textbf{Sabin Huda:} Conceptualization, Resources, Methodology, Investigation, Formal analysis, Visualization, Software, Writing – original draft, Writing – review and editing. \textbf{Ernest Foo:} Conceptualization, Writing – review and editing, Supervision, Validation. \textbf{Zahra Jadidi:} Writing – review and editing, Supervision. \textbf{M A Hakim Newton:} Writing – review and editing, Supervision. \textbf{Abdul Sattar:} Project administration, Supervision.

\section*{Declaration of Competing Interest}
The authors declare that they have no known competing financial interests or personal relationships that could have appeared to influence the work reported in this article.

\section*{Data and Code Availability Statement}
The AMLNet synthetic transaction dataset (Version 1.0; 1,090,173 records; approx 0.16\% laundering-labelled) underlying all reported analyses is publicly available at Zenodo: DOI: \url{https://doi.org/10.5281/zenodo.16736515} under a CC BY-NC 4.0 license. A synchronized copy (mirror) will be added to Kaggle for discoverability. For citation, please reference the dataset DOI (Version 1.0) and the accompanying manuscript as outlined in the Zenodo record description. The framework source code (data generator and detection scripts) will be released in a public GitHub repository at (or prior to) publication; core algorithmic details necessary for reproducibility are included in the manuscript.

\section*{Responsible Use and Limitations}
AMLNet is a fully synthetic and regulation-aware resource for AML research and benchmarking; it contains no real customer data and is designed to accelerate transparent experimentation, while any production compliance deployment requires institution-specific validation, risk assessment, and governance. The 75\% regulatory alignment score reflects systematic coverage of a curated, publicly interpretable subset of AUSTRAC rules, and the 0.75 technical fidelity score evidences balanced temporal, structural, and behavioral realism. The prevalence of laundering positives is intentionally low (approximately 0.16\%) yet may still exceed rates in some institutions; select rare typologies, jurisdictional nuances, and threshold dynamics are simplified to reduce misuse and re-identification risk. Users must independently verify suitability under local AML/CTF and data protection regulations and should avoid employing the resource for adversarial evasion scenario design without appropriate oversight. The dataset is released under CC BY-NC 4.0; any adaptation for operational compliance contexts requires additional validation and governance controls.

\section*{Declaration of Generative AI and AI-assisted Technologies}
During the preparation of this work, the author(s) used generative AI tools, including ChatGPT and Claude, to enhance technical writing. All content generated by these tools was subsequently reviewed and edited by the author(s), who take full responsibility for the content.

\section*{Suggested Citation}
Huda, S., Foo, E., Jadidi, Z., Newton, M. A. H., \& Sattar, A. (2025). AMLNet: A Knowledge-Based Multi-Agent Framework to Generate and Detect Realistic Money Laundering Transactions. Preprint (Version 1, 15 Sep 2025). Under review in Expert Systems with Applications. Dataset: \url{https://doi.org/10.5281/zenodo.16736515}.

\bibliographystyle{elsarticle-harv}
\bibliography{main}

\end{document}